\documentclass[11pt]{article}
\usepackage{acl}
\usepackage{times}
\usepackage{latexsym}
\usepackage[T1]{fontenc}
\usepackage[utf8]{inputenc}
\usepackage{microtype}
\usepackage{inconsolata}
\usepackage{graphicx}
\usepackage{booktabs}
\usepackage{multirow}
\usepackage{colortbl}
\usepackage{xcolor}
\usepackage{array}
\usepackage{amsmath}
\usepackage{amssymb}
\usepackage{amsfonts}
\usepackage{bm}
\usepackage{makecell}
\usepackage{tabularx}
\usepackage{placeins}
\usepackage{dblfloatfix}
\newcolumntype{Y}{>{\raggedright\arraybackslash}X}

\title{A Pinch of SFT, A Dash of RL:\\When Reinforcement Learning Helps Long-Horizon Advertising Agents}

\author{
\textbf{Aakash Kolekar}$^{1}$,
\textbf{Sahika Genc}$^{2}$,
\textbf{Bunyamin Sisman}$^{1}$,
\textbf{Shahriar Shariat}$^{1}$, \\
\textbf{Shree Vandana Kachroo}$^{1}$,
\textbf{Avishek Saha}$^{1}$,
\textbf{Qianli Wu}$^{1}$,
\textbf{Ari Singer}$^{1}$,
\textbf{Benoit Dumoulin}$^{1}$ \\
$^{1}$Amazon Advertising, $^{2}$Amazon Web Services Agentic AI\\
\{\texttt{aakashvv,sahika,bunyamis,sshariat,kacshree,avisaha,qianliwu,arising,bdumouli}\}\texttt{@amazon.com}
}
\begin{document}
\maketitle

\begin{abstract}
Enterprise analytics agents solve long-horizon tool-use problems over distributed business data, requiring retrieval, reasoning, API calls, code execution, and adaptation to intermediate observations. Supervised fine-tuning (SFT) calibrates tool syntax and teacher-supported behavior, whereas reinforcement learning (RL) can explore reward-supported behaviors beyond demonstrations; applied uniformly, however, RL can perturb already-calibrated skills. We study how to balance SFT and RL under production-mirroring beta APIs. We observe that, in our controlled experiment, checkpoint trajectories retrospectively separated into three regimes: \emph{Imitation}, where SFT captured reliable teacher behavior; \emph{Lift}, where both stages helped; and \emph{Discovery}, where useful reward-observable behavior lay outside reliable teacher support. We leverage this prospectively, using teacher support and reward-observable headroom to route features to SFT only, SFT$\rightarrow$RL, increased RL allocation, or further environment development. Across 18 subsequent feature-specific experiments, the diagnostic predicted 15/18 observed trajectories. On GPT-OSS 120B, targeted SFT$\rightarrow$RL produced positive point estimates on 7/8 advertiser skills relative to a frontier Control; five positive gains had paired 95\% confidence intervals excluding zero, while one skill had a confidence-supported regression. The largest gain was non-disclosure ($+11.27$ points; 95\% CI $[+9.72,+12.82]$). A separate SME audit surfaced that targeted RL reduces standard leakage from 11.8\% to 2.9\% and adversarial leakage from 22.9\% to 6.8\% relative to SFT while preserving actionability (86.2\% to 85.7\%). In a matched uniform-versus-targeted comparison with shared rewards and optimization, targeted RL improved the seven-skill mean delta from $+1.62$ to $+3.57$ while using 43\% less incremental RL compute.
\end{abstract}

\begin{figure*}[t]
\centering
\includegraphics[width=\textwidth]{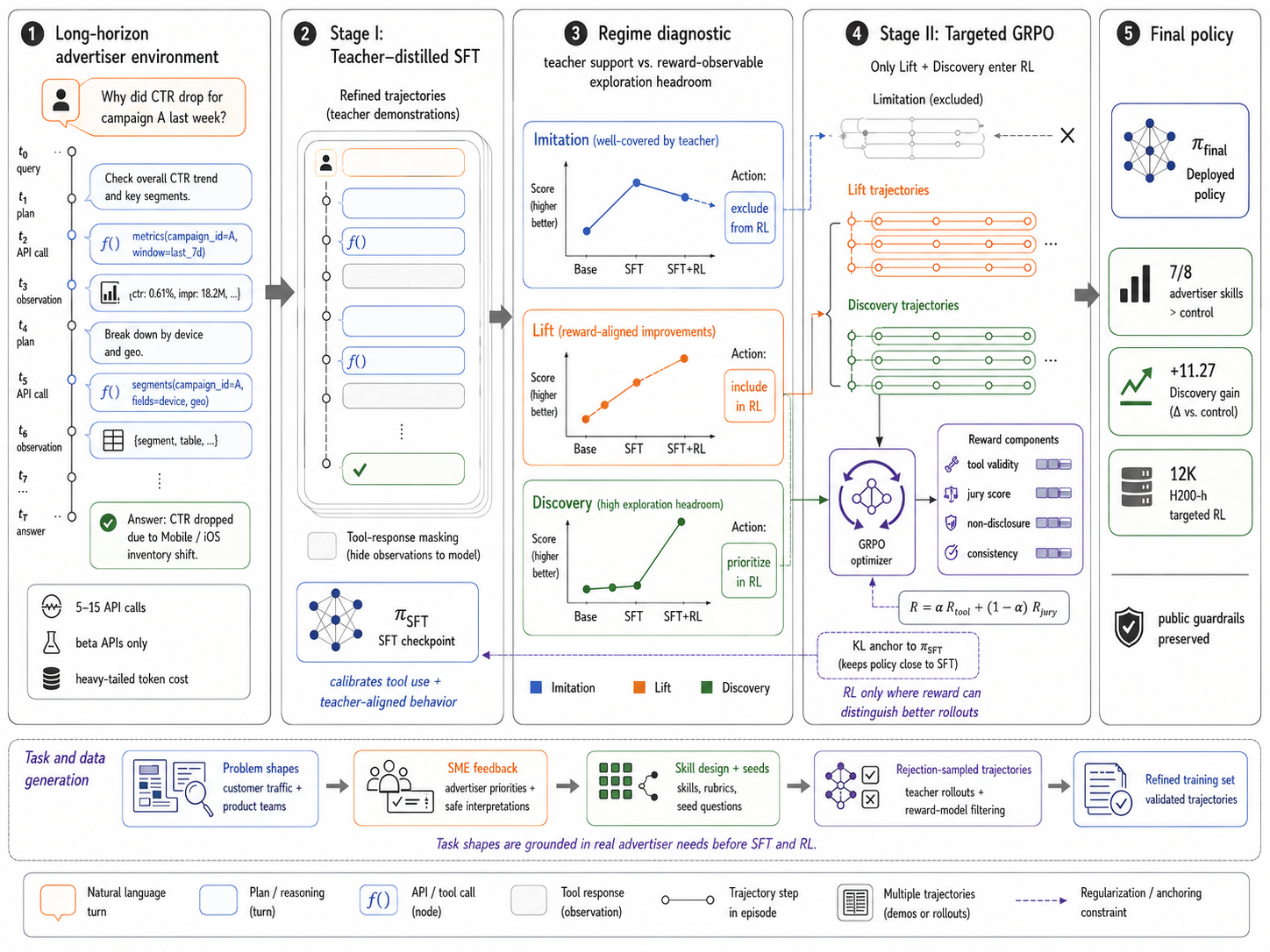}
\caption{\textbf{Targeted post-training pipeline.}
Episodes interleave reasoning, beta-API calls, and tool observations. Stage~I uses teacher-distilled SFT with tool-response masking. The original eight-skill study motivates a three-regime hypothesis; a subsequent prospective diagnostic uses teacher support and frozen-SFT reward headroom to route new features before RL. Stage~II applies GRPO only where reward-observable headroom justifies on-policy optimization, using tool-validity and skill-aware jury rewards with a KL anchor to $\pi_{\mathrm{SFT}}$.}
\label{fig:method_overview}
\end{figure*}

\section{Introduction}
\label{sec:intro}

\noindent An advertiser running a holiday campaign across hundreds of audience segments does not ask a single, well-scoped question. They ask \emph{``why is my return on ad spend down 12\% this week, which audiences are driving the drop, and where should I move my budget?''} Answering this requires identifying affected campaigns, retrieving metrics across a date window, decomposing the drop by segment, and recommending spend shifts that respect advertiser objectives. Such requests are naturally long-horizon tool-use problems: success is defined by outcomes over a sequence of API calls, code execution steps, and intermediate observations, not by a single next-token target. This makes reinforcement learning attractive, but not uniformly beneficial. Pure RL from a cold-started base model collapses on long-horizon advertiser skills: when nearly every rollout fails, the group-relative advantages that drive GRPO updates \citep{shao2024deepseekmath} contain little usable signal.

\noindent The central question is how to allocate SFT and RL under limited rollout budget without broad capability regression. Teacher-distilled SFT calibrates tool syntax and common reasoning traces, but can plateau when demonstrations do not reliably cover desired behavior. Post-SFT RL therefore poses a sharper allocation problem: \emph{when does on-policy optimization discover reward-supported behavior beyond the teacher, and when does it merely perturb behavior SFT has already calibrated?}

\noindent In the original controlled experiment, we observe three retrospective training trajectories. In \emph{Imitation}, demonstrations are reliable and SFT closes most of the gap. In \emph{Lift}, SFT establishes a competent policy and RL adds further improvement. In \emph{Discovery}, the desired behavior is reward-observable but weakly represented in teacher demonstrations. We subsequently operationalize this hypothesis into a pre-RL diagnostic using teacher support and reward-observable headroom. The diagnostic determines whether a feature receives SFT only, SFT followed by RL, increased RL allocation, or additional environment (RL Gym) development. This separates the retrospective evidence that motivated the taxonomy from prospective evidence about whether it can guide new experiments.

\noindent Our contributions are:
\textbf{(i) A regime hypothesis and prospective diagnostic:} we operationalize the three regimes using teacher support and reward-observable headroom measured before feature-specific RL; across 18 subsequent experiments, the diagnostic predicts 15/18 observed cases.
\textbf{(ii) A targeted post-training recipe with a matched comparison:} teacher-distilled, tool-response-masked SFT is followed by GRPO only where exploration is reward-distinguishable. Under matched rewards and optimization settings, targeted allocation improves the seven-skill mean delta from $+1.62$ to $+3.57$ and uses 43\% less incremental RL compute than uniform RL.
\textbf{(iii) SME-grounded evaluation with paired uncertainty and human safety validation:} every headline score is Avg@32 on fixed held-out examples; we report paired 95\% intervals and separately audit non-disclosure with human SMEs, showing substantially lower leakage without a material loss in actionability.
\textbf{(iv) Production-scale evidence and accounting:} we report failed and successful large-scale runs, explicit offline release-readiness criteria, complete per-recipe SFT/RL compute, and the systems modifications required for long-context MoE RL under production-mirroring APIs.

\section{The Advertiser Environment}
\label{sec:method}

\noindent We instantiate the beta-API infrastructure as a multi-turn tool-calling RL gym. At step $t$, the state $s_t$ contains the user request, system instructions, prior reasoning, tool calls, and returned observations. The action $a_t$ is the next assistant turn: either a structured call to one of 15 beta APIs, optionally preceded by planning tokens, or a terminal response. The environment executes valid tool calls, appends observations to the trajectory, and scores terminal responses with $R(\tau)$. The policy maximizes $\mathbb{E}_{\tau\sim\pi_\theta,P}[R(\tau)]$. Both policy and environment contribute variance: two rollouts from the same prompt can differ in the calls sampled, in the observations returned, or both. The prompt-length distribution is heavy-tailed, with a 99th percentile of 116{,}445 tokens; rollout-cost details are moved to Appendix~\ref{app:cost_figures}.

\subsection{Task and Data Generation}
\label{sec:data_generation}

\noindent We source \emph{problem shapes} from anonymized customer-traffic patterns, product-team workflow requests, and recurring analytics intents observed by advertiser-facing teams. Subject-matter experts (SMEs) review these shapes for business relevance, policy sensitivity, and advertiser-persona coverage; their feedback defines the skills in Appendix~\ref{app:task_taxonomy}, seed questions, API surfaces, and rubric dimensions. Teacher models then generate candidate multi-turn trajectories against beta APIs. We retain only trajectories that pass execution validation, data-integrity checks for metric and entity consistency, and a judge-based reasoning audit for plan--code alignment, grounding, and skill-specific rubric satisfaction. Rejections feed back into prompt and rubric refinement; accepted trajectories form the masked-SFT corpus, and the same metadata is retained for regime diagnosis and reward routing.

\subsection{Trajectory Format and SFT}
\label{sec:sft}

\noindent A teacher produces multi-turn trajectories whose blocks alternate \texttt{<think>}, \texttt{<tool\_call>}, \texttt{<tool\_response>}, and \texttt{<final\_response>}. We train with selective loss masking over assistant-generated tokens, following standard practice for multi-turn tool-use trajectories \citep{qin2024toolllm}, normalizing by the number of unmasked tokens so long tool-heavy trajectories do not dominate the gradient by length. Loss is applied to \texttt{<think>}, \texttt{<tool\_call>}, and \texttt{<final\_response>} tokens, and masked on tool responses, system, and user tokens. We include Toucan-1.5M trajectories \citep{xu2025toucan} as an out-of-domain regularizer.

\paragraph{Why the two stages differ.}
For skill $k$, masked teacher-distilled SFT is equivalent up to constants to minimizing a forward divergence from the demonstrated distribution $p_{\mathrm{teach}}^k$ to the policy,
\begin{equation}
\mathcal{L}_{\mathrm{SFT}}^k \equiv D_{\mathrm{KL}}\!\left(p_{\mathrm{teach}}^k \,\|\, \pi_\theta\right) + \mathrm{const.}
\label{eq:sft_forward_kl}
\end{equation}
This is appropriate when the teacher reliably covers the desired behavior, but it cannot directly reward successful behaviors absent from demonstrations. The KL-regularized reward-maximization objective instead has the reward-tilted solution
\begin{equation}
\pi_k^{*}(\tau) \propto \pi_{\mathrm{SFT}}(\tau)\,\exp\!\left(R_k(\tau)/\beta\right),
\label{eq:reward_tilted_policy}
\end{equation}
which preserves the SFT policy as a reference while shifting mass toward trajectories the environment scores highly.

\subsection{Prospective Regime Diagnostic}
\label{sec:prospective_diagnostic}

\noindent The eight-skill decomposition in the controlled study is retrospective. To make the hypothesis actionable for new features, we subsequently define two quantities measured \emph{before} feature-specific RL. Let $\mathrm{Valid}(\tau)$ indicate that a teacher trajectory passes execution, data-integrity, and SME-authored rubric checks. Teacher support and reward-observable headroom are
\begin{align}
T_k &= \Pr_{\tau\sim p^k_{\mathrm{teacher}}}\!\left[\mathrm{Valid}(\tau)=1\right],
\label{eq:teacher_support}\\[-1mm]
H_k &= \mathbb{E}_{q}\!\left[\max_{i\leq G}R(\tau_i)-\frac{1}{G}\sum_{i=1}^{G}R(\tau_i)\right],\quad G=8,
\label{eq:headroom}
\end{align}
where the $G$ trajectories are sampled from the frozen SFT policy for each prompt. We route features before RL using the following rule:
\begin{center}
\footnotesize
\setlength{\tabcolsep}{3.2pt}
\renewcommand{\arraystretch}{1.08}
\begin{tabularx}{\columnwidth}{@{}c c >{\raggedright\arraybackslash}X@{}}
\toprule
\textbf{Teacher support} & \textbf{Reward headroom} & \textbf{Route} \\
\midrule
$T_k\geq0.80$ & $H_k<0.10$ & Imitation: SFT only \\
$T_k\geq0.80$ & $H_k\geq0.10$ & Lift: SFT$\rightarrow$RL \\
$T_k<0.80$ & $H_k\geq0.10$ & Discovery: increased RL \\
$T_k<0.80$ & $H_k<0.10$ & Improve environment \\
\bottomrule
\end{tabularx}
\end{center}
Assignments are fixed before each feature-specific RL experiment; Section~\ref{sec:prospective_results} reports the resulting 18-feature evaluation.

\subsection{Targeted GRPO with a Jury Reward}
\label{sec:rl}

\noindent We refine the SFT checkpoint with GRPO \citep{shao2024deepseekmath} against the 15-tool beta environment. For each prompt $q$, we sample a group of $G$ trajectories and compute a group-normalized advantage:
\begin{equation}
\hat{A}_i = \frac{R(\tau_i) - \frac{1}{G}\sum_{j=1}^{G} R(\tau_j)}{\mathrm{std}(\{R(\tau_j)\}_{j=1}^{G}) + \epsilon}.
\label{eq:advantage}
\end{equation}
The reward combines tool grounding and semantic quality,
\begin{equation}
R(\tau) = \alpha\,R_{\mathrm{tool}}(\tau) + (1-\alpha)\,R_{\mathrm{jury}}(\tau),
\label{eq:reward}
\end{equation}
where $R_{\mathrm{tool}}\in[0,1]$ penalizes invalid calls, unsupported joins, and execution failures, and $R_{\mathrm{jury}}\in[0,1]$ scores semantic answer quality. The tool-grounding term reduces the chance that the policy learns fluent but unsupported responses.

\paragraph{Skill-aware jury reward.}
$R_{\mathrm{jury}}$ is skill-aware rather than a universal rubric. Every trajectory is scored for thought consistency, response factuality, and tool-output validity; skill-specific dimensions such as numerical sanity, audience analysis, performance-gap attribution, and non-disclosure are activated only when relevant (Appendix~\ref{app:cost_figures}). The non-disclosure rubric independently scores restricted-information leakage, factual correctness, actionability, and excessive abstraction, so a vague refusal cannot score highly merely by omitting sensitive values. The reward jury combines open-weight judges from different model families---DeepSeek-R1 and Qwen3-235B-A22B---to reduce self-preference bias \citep{guo2025deepseekr1,qwen3_2025,panickssery2024llm,zheng2023judging,verga2024replacing}. On a 200-trajectory human-rated calibration set, ensemble jury--human Spearman is 0.78 and two SME raters achieve Cohen's $\kappa=0.74$ (Appendix~\ref{app:jury}).

\paragraph{Stabilization and infrastructure.}
Disaggregated SGLang \citep{sglang2024} rollouts and Megatron \citep{shoeybi2019megatron} training introduce rollout--training mismatch even with periodic synchronization \citep{yao2025rolloutmismatch,zheng2025stabilizing}. We use a KL anchor to $\pi_{\mathrm{SFT}}$, asymmetric policy-ratio clipping, and token-level truncated importance sampling (TIS), tracking held-out reward, entropy, clipping rate, and TIS truncation. GPT-OSS-120B \citep{gptoss2025} training additionally required bit-correct MoE weight synchronization, context-parallel support for learnable-softmax attention, and token-budget batching for heavy-tail trajectories. Full objectives and systems details are in Appendices~\ref{app:optimization_diagnostics} and \ref{app:infra}.

\begin{table*}[t]
\caption{\textbf{Per-skill evaluation against Control on GPT-OSS 120B.} Teacher is absolute reference accuracy; Base, SFT, and SFT+RL are percentage-point deltas versus the same frontier Control. Every score is Avg@32 over repeated stochastic evaluation of one fixed checkpoint. The final column reports paired 95\% CIs for SFT+RL. Bold deltas have intervals excluding zero.}
\centering
\footnotesize
\setlength{\tabcolsep}{4.2pt}
\renewcommand{\arraystretch}{1.08}
\begin{tabularx}{\textwidth}{@{}>{\raggedright\arraybackslash}X@{\hspace{7pt}}rrrrr>{\centering\arraybackslash}p{0.18\textwidth}@{}}
\toprule
\textbf{Skill} & \textit{Teacher} & \textbf{Control} & \textbf{Base $\Delta$} & \textbf{SFT $\Delta$} & \textbf{SFT+RL $\Delta$} & \textbf{95\% paired CI} \\
\midrule
Measure funnel & 94.32 & 91.73 & $-2.73$ & $+1.55$ & $+0.41$ & $[-0.28,+1.10]$ \\
Target / audience strategy & 90.92 & 90.00 & -- & -- & $\bm{+1.50}$ & $[+0.48,+2.52]$ \\
Business metrics & 94.42 & 91.22 & $+4.78$ & $+3.19$ & $\bm{+5.18}$ & $[+4.03,+6.33]$ \\
Categorize campaigns / ads & 59.19 & 57.05 & $-6.95$ & $-4.35$ & $\bm{-2.05}$ & $[-3.21,-0.89]$ \\
Campaign performance & 84.70 & 82.51 & $+5.19$ & $+5.49$ & $\bm{+7.49}$ & $[+6.15,+8.83]$ \\
Identify product & 89.89 & 87.56 & $-2.34$ & $-1.13$ & $+0.66$ & $[-0.14,+1.46]$ \\
Brand voice & 96.65 & 96.67 & $-2.82$ & $+1.15$ & $\bm{+2.03}$ & $[+1.07,+2.99]$ \\
Share insights without disclosing non-public information & 84.53 & 84.23 & $-1.00$ & $-0.23$ & $\bm{+11.27}$ & $[+9.72,+12.82]$ \\
\bottomrule
\end{tabularx}
\label{tab:main}
\end{table*}

\section{Results}
\label{sec:results}

\paragraph{Evaluation protocol and baselines.}
Each domain skill contains at least 250 held-out examples; exact counts are confidential. All checkpoints and Control are evaluated on identical fixed examples with no overlap with SFT or RL data. Every reported score is Avg@32 over 32 independent stochastic evaluation repetitions of the \emph{same trained checkpoint}; these repetitions capture decoding and beta-environment stochasticity and are not 32 independently trained policies. We report paired 95\% confidence intervals from the matched evaluation outputs following a paired resampling protocol \citep{koehn2004bootstrap}.

Teacher and Control are frontier closed-weight models with at least 256K context, the same 15 beta APIs, and full production-style schemas. Teacher generates candidate SFT trajectories under fixed low-temperature decoding; Control is the fixed production comparison baseline. Neither is exposed to candidate post-training data, participates in the RL reward, or participates in the Kimi-K2 headline evaluation. Candidate policies and Control use the same fixed prompts, tool interfaces, and beta-environment configuration. Headline deltas are scored by Kimi-K2 \citep{kimik2_2025}, from a different model family than the DeepSeek-R1/Qwen3 reward jury.

\paragraph{Controlled eight-skill result.}
In Table~\ref{tab:main} targeted SFT+RL has positive point estimates on 7/8 skills versus Control; five gains exceed one percentage point and have paired intervals excluding zero. Measure Funnel ($+0.41$) and Identify Product ($+0.66$) are directional because their intervals cross zero. Categorize Campaigns/Ads significantly regresses by $-2.05$ points. Its teacher reference is the weakest in the suite (59.19 versus 84.53 next-lowest), and deltas improve monotonically without closing the gap. The matched seven-skill subset, scored at Base, SFT, and SFT+RL, moves from 2/7 to 4/7 to 6/7 positive point estimates, with four SFT+RL gains above one point. Public guardrails remain non-negative for SFT+RL: IFEval \citep{zhou2023ifeval} $+1.22$, GSM8K \citep{cobbe2021gsm8k} $+3.43$, and GPQA-Diamond \citep{rein2024gpqa} $+7.07$ points versus Control.

\paragraph{Retrospective three-regime decomposition.}
\label{sec:decomp}

\begin{figure}[t]
\centering
\includegraphics[width=\columnwidth]{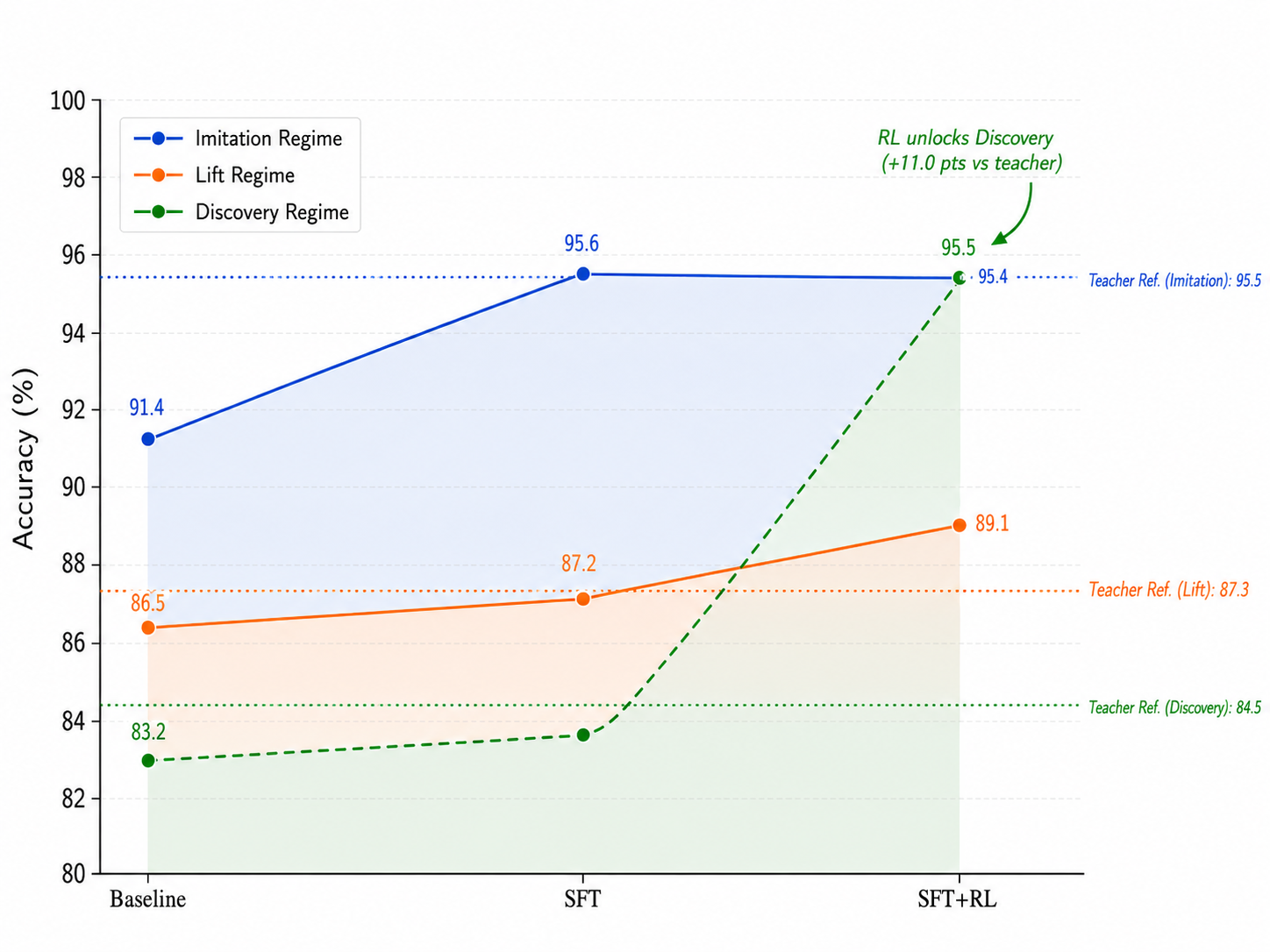}
\caption{\textbf{Retrospective three-regime decomposition.} Mean accuracy for representative GPT-OSS 120B skills across Base$\rightarrow$SFT$\rightarrow$targeted RL. Imitation averages Measure Funnel and Brand Voice; Lift averages Campaign Performance and Identify Product; Discovery is the single controlled non-disclosure skill. Dotted lines denote teacher references. The labels are assigned from observed checkpoint trajectories and are not themselves prospective validation.}
\label{fig:decomp}
\end{figure}

Figure~\ref{fig:decomp} summarizes the original retrospective observation. Imitation moves 91.43$\rightarrow$95.55$\rightarrow$95.42; Lift 86.46$\rightarrow$87.22$\rightarrow$89.11; and Discovery 83.23$\rightarrow$84.00$\rightarrow$95.50, exceeding the teacher by 10.97 points. Discovery contains one controlled safety-critical skill, so it is an existence result rather than evidence of regime prevalence.

\begin{figure*}[!t]
\centering
\vspace{-2mm}
\includegraphics[width=0.96\textwidth,trim=8 8 8 8,clip]{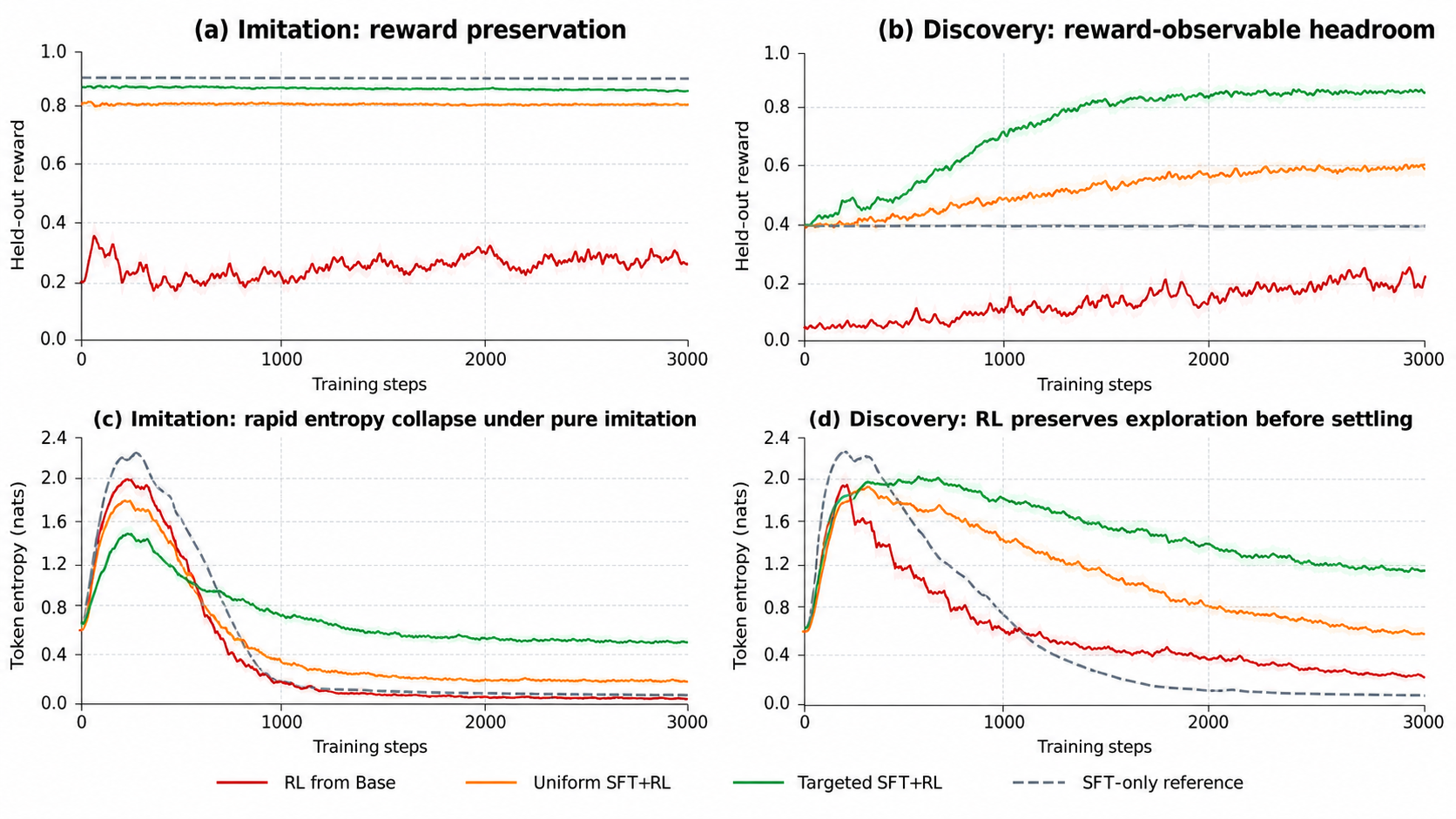}
\vspace{-1mm}
\caption{\textbf{Regime-conditioned reward and entropy dynamics.} Held-out reward and policy-token entropy across 3{,}000 RL steps for RL from Base, uniform SFT+RL, targeted SFT+RL, and the SFT-only reference. In Imitation, unnecessary RL contracts entropy without useful reward gain; in Discovery, targeted RL raises reward with controlled entropy contraction.}
\vspace{-2mm}
\label{fig:reward_entropy_diagnostics}
\end{figure*}

\paragraph{Regime-conditioned training dynamics.}
Figure~\ref{fig:reward_entropy_diagnostics} complements the endpoint decomposition. The SFT reference already occupies a high-reward region for Imitation, and targeted allocation preserves it by excluding that regime from RL; uniform optimization instead reduces entropy without improving reward. In Discovery, targeted RL produces sustained held-out reward gains while entropy contracts gradually rather than collapsing. Update-magnitude, policy-clipping, and TIS diagnostics remain in Appendix~\ref{app:optimization_diagnostics}.

\paragraph{Matched uniform versus targeted RL.}
To isolate the value of allocation more directly, we compare retained uniform and targeted runs that share the SFT checkpoint, reward jury and tool reward, optimizer and learning rate, KL coefficient, clipping/TIS configuration, and rollout group size. The primary difference is the skill-sampling curriculum: uniform RL samples all skills, while targeted RL excludes Imitation and overweights Discovery.

\begin{table}[t]
\centering
\small
\setlength{\tabcolsep}{4pt}
\begin{tabular}{@{}lcc@{}}
\toprule
\textbf{Matched seven-skill result} & \textbf{Uniform} & \textbf{Targeted} \\
\midrule
Positive vs. Control & 4/7 & \textbf{6/7} \\
Mean $\Delta$ vs. Control & $+1.62$ & $\bm{+3.57}$ \\
Imitation mean accuracy & 94.31 & \textbf{95.42} \\
Lift mean accuracy & 88.34 & \textbf{89.11} \\
Discovery accuracy & 90.72 & \textbf{95.50} \\
Incremental RL compute & 21K & \textbf{12K} \\
Total SFT+RL compute & 36K & \textbf{27K} \\
\bottomrule
\end{tabular}
\caption{Matched uniform-versus-targeted allocation. Compute is in H200-hours. Targeted allocation uses 43\% less incremental RL compute.}
\label{tab:uniform_targeted}
\end{table}

Targeted allocation improves Discovery, better preserves the SFT-calibrated Imitation mean, and uses 43\% less incremental RL compute. Development sweeps not retained in this matched comparison sometimes differed along additional dimensions; we therefore interpret Table~\ref{tab:uniform_targeted} as evidence for the matched allocation comparison and avoid attributing every recipe-level gain solely to routing. A sanitized qualitative contrast between the SFT and targeted-RL responses is given in Appendix~\ref{sec:case_studies}.

\paragraph{Prospective operational evidence.}
\label{sec:prospective_results}
The diagnostic in Section~\ref{sec:prospective_diagnostic} was subsequently applied before feature-specific RL across multiple business lines. It predicted the observed post-training trajectory for 5/6 Imitation assignments, 6/7 Lift assignments, and 4/5 Discovery assignments: 15/18 (83.3\%) overall. This follow-on study is broader but less controlled than the original eight-skill experiment; it supports the diagnostic's operational usefulness rather than establishing a universal taxonomy. Low-support/low-headroom cases are returned for additional SME annotation, reward improvement, or environment development. Further operational details are in Appendix~\ref{app:prospective_details}.

\paragraph{Independent human audit of non-disclosure.}
The largest controlled effect is non-disclosure ($+11.27$ points, 95\% CI $[+9.72,+12.82]$). To test whether this reflects genuine abstraction rather than generic vagueness, a separate held-out set is independently audited by SMEs using standard and adversarial prompts. The rubric separately measures leakage, correctness, actionability, refusal, and excessive abstraction; inter-annotator agreement is Cohen's $\kappa=0.76$, distinct from the $\kappa=0.74$ jury-calibration set.

\begin{table}[t]
\centering
\small
\setlength{\tabcolsep}{4.0pt}
\begin{tabular}{@{}lccc@{}}
\toprule
\textbf{Human-audit metric} & \textbf{Control} & \textbf{SFT} & \textbf{SFT+RL} \\
\midrule
Standard leakage & 12.4\% & 11.8\% & \textbf{2.9\%} \\
Adversarial leakage & 24.7\% & 22.9\% & \textbf{6.8\%} \\
Informative/actionable & 84.6\% & 86.2\% & 85.7\% \\
Acceptably safe & 76.9\% & 78.4\% & \textbf{91.8\%} \\
\bottomrule
\end{tabular}
\caption{Independent SME audit of the non-disclosure checkpoint. Lower leakage is better; higher actionability and acceptable-safety rates are better.}
\label{tab:human_audit_main}
\end{table}

Relative to SFT, targeted RL reduces standard leakage by 8.9 points and adversarial leakage by 16.1 points while changing actionability by only $-0.5$ points. This human audit, rather than jury correlation alone, is the primary evidence that the gain is not produced by generic refusal or excessive abstraction. It remains a held-out offline validation, not a complete real-world safety certification.

\paragraph{Offline release readiness, compute, and serving-period checks.}
The pre-defined gate requires mean domain delta $\geq+2.0$, at least 6/8 positive domain point estimates, no individual domain regression below $-3.0$, non-disclosure gain over Teacher $\geq+5.0$, zero negative public guardrails, jury--human Spearman $\rho\geq0.75$, tool-call validity $\geq97\%$, P95 latency regression no worse than $+10\%$, and standard-prompt leakage below 5\%. Targeted SFT+RL passes all nine criteria (Appendix~\ref{app:release_gate}). The shared SFT stage costs 15K H200-hours; uniform RL adds 21K (36K total) and targeted RL adds 12K (27K total). RL from Base does not pass the gate within 25K H200-hours.

The same candidate checkpoint was also measured against Control during a fixed online evaluation period using the same traffic slice and serving configuration. It records $+5.4$ percentage points in multi-turn task completion, $+7.1$ points in recovery after a failed tool call, $+3.6$ points in recommendation acceptance, and a $-6.8\%$ change in P95 user-perceived latency; the full operational table is in Appendix~\ref{app:production_metrics}. Confidentiality requirements prevent disclosure of the exact evaluation window, traffic volume, and whether individual measurements were collected in shadow or live mode. We therefore treat these results as operational evidence that the candidate does not require an obvious serving-performance trade-off, not as a reproducible randomized causal estimate of online product impact.

\section{Related Work}
\label{sec:related_work}

\paragraph{SFT, RL, generalization, and retention.}
Post-training builds on preference-based RL and RLHF \citep{christiano2017deep,ziegler2019fine,stiennon2020learning,ouyang2022instructgpt,bai2022constitutional,rafailov2023dpo}. Recent work directly contrasts SFT and RL: \citet{chu2025sftgeneralizes} find that SFT tends to memorize while outcome-based RL can generalize to unseen variants; \citet{shenfeld2026razor} argue that on-policy RL is biased toward lower-KL solutions that retain prior capabilities; and \citet{chen2026retaining} isolate on-policy data as an important mechanism for mitigating forgetting. Two-stage SFT$\rightarrow$RL efficiency has also been observed in mathematical reasoning \citep{yoshihara2025practicaltwostagerecipemathematical}. Our question is complementary: within one heterogeneous long-horizon agent, can observable pre-RL properties determine \emph{where} RL budget should be allocated? Complementary lines of work change \emph{what} the policy trains on rather than where budget is spent: curriculum scheduling from easy to hard tasks \citep{parashar2025curriculum}, and mixing off-policy teacher traces into on-policy rollouts \citep{yan2025luffy}. We additionally connect reasoning-RL systems \citep{shao2024deepseekmath,guo2025deepseekr1,lambert2024tulu3} with reward over-optimization \citep{gao2023scaling,skalse2022defining} and judge calibration \citep{lambert2025rewardbench,zheng2023judging,panickssery2024llm,verga2024replacing}.

\paragraph{Tool use and industrial e-commerce systems.}
Tool-use and function-calling work \citep{schick2023toolformer,li2023apibank,patil2024gorilla,qin2024toolllm,patil2025bfcl,tang2023toolalpaca} and agent benchmarks \citep{yao2023react,liu2024agentbench,zhou2024webarena,yao2025taubench} emphasize interleaved reasoning and action. Industrial e-commerce systems have separately used retrieval-augmented generation and domain alignment for product-aware query completion and conversational shopping \citep{sun2024productrag,luo2025shopping}. Selective-information generation has also been studied in blocked textual-graph QA, where relevant information must be inferred yet excluded from the answer \citep{yan2025taona}; our non-disclosure skill differs in operating over tool-returned enterprise observations and learning the behavior through targeted post-SFT RL. Efficient LLM-RL systems separate rollout and training backends, creating an implicit off-policy gap that motivates TIS and related corrections \citep{yao2025rolloutmismatch,zheng2025stabilizing}. Imitation-regime regressions are also consistent with catastrophic interference \citep{mccloskey1989catastrophic}, alignment-induced generalization shifts \citep{kirk2024understanding}, and long-context degradation \citep{liu2024lost}.

\section{Discussion and Conclusion}
\label{sec:discussion}

The controlled study supports a stage-wise allocation view of post-training. SFT is effective when teacher support is high, while RL is most useful when the frozen SFT policy exposes reward-distinguishable alternatives. The matched comparison strengthens this: targeted allocation improves the seven-skill mean delta while reducing incremental RL compute, rather than applying more optimization everywhere. The prospective diagnostic converts that observation into an actionable routing rule.

The evidence has an important asymmetry. Imitation and Lift are supported by multiple controlled skills, whereas Discovery is represented by one controlled proprietary non-disclosure skill. The 4/5 prospective Discovery result broadens the operational evidence but is not a substitute for a larger controlled Discovery benchmark. Similarly, the Qwen3-32B run (Appendix~\ref{app:model_transfer}) shows qualitative Discovery transfer but weaker preservation elsewhere, indicating that KL strength, reward mixture, and rollout allocation do not automatically transfer across scale and architecture.

Finally, the non-disclosure result illustrates why semantic reward design requires independent human validation. The model does not merely refuse more often: standard and adversarial leakage drop sharply while actionability remains nearly unchanged. Together, these results suggest a practical recipe for long-horizon enterprise agents: use SFT to establish domain competence, measure teacher support and reward-observable headroom before spending RL budget, apply RL selectively where exploration is distinguishable, and validate safety-critical gains outside the reward model.

\FloatBarrier
\section*{Limitations}

The proprietary advertiser benchmark contains eight controlled domain skills, and the Discovery aggregate contains one safety-critical skill. The retrospective three-regime decomposition should therefore be read as an empirical hypothesis supported by observed checkpoint trajectories, not a population-level taxonomy. The later prospective analysis covers 18 feature experiments and predicts 15/18 cases, but remains modest in scale and is less controlled than the original experiment. Broadening controlled Discovery evidence remains the most important validation step; public BFCL/ToolBench results in Appendix~\ref{app:public_details} are post-hoc structural analogues, not external validation.

Exact per-skill prompt counts cannot be disclosed, although each skill contains at least 250 held-out examples and the largest is less than $1.4\times$ the smallest. Avg@32 captures stochastic evaluation and beta-environment variance for one trained checkpoint; it does not estimate variance across independently trained policies. The paired intervals therefore quantify evaluation uncertainty conditional on the retained checkpoints, not end-to-end training-seed uncertainty.

The independent non-disclosure audit substantially strengthens the safety evidence, but it remains a held-out offline audit rather than a complete real-world safety certification. Learned abstraction may still fail under distribution shift, new adversarial attacks, or changes in tool outputs, so deployment requires least-privilege access, output-side controls, logging, and ongoing review. Operational metrics are also limited by confidentiality: the exact evaluation window, traffic volume, and shadow/live assignment mode cannot be disclosed, preventing full reproduction or a clean causal interpretation of those serving-period changes.

Production data and beta APIs are proprietary, and the exact hyperparameter sensitivity surface is not released. Recipe-level compute includes SFT and RL costs attributable to each retained training recipe; shared development costs from hyperparameter search and infrastructure bring-up are excluded from those recipe totals because they are not cleanly attributable to one recipe. Finally, the environment uses relatively stable beta-API interfaces; rapidly changing schemas would require periodic trajectory refresh, regression evaluation, and targeted re-training.

\section*{Ethical Considerations}

The model operates over advertiser data accessed through beta APIs that mirror production, under advertiser-isolation and access-control constraints. Tool observations may contain sensitive or competitor-derived values even when the final response should expose only abstracted business guidance. Non-disclosure is therefore a behavioral guardrail, not an information barrier. Any production use must combine model behavior with least-privilege tool access, output-side policy filters, audit logging, and independent review of sensitive-output scenarios. The non-disclosure audit reported here uses SME-authored rubrics and a held-out annotation set, but does not eliminate risks under distribution shift or adversarial prompting. The reward jury is open-weight and inspectable; teacher-generated trajectories are used only under the applicable provider terms and internal approvals for model-development and distillation workflows.

\bibliographystyle{acl_natbib}
\bibliography{custom}

\clearpage
\onecolumn
\appendix

\section{Regime Aggregates and Complete Recipe Compute}
\label{app:regime_details}

\begin{table}[!hbt]
\centering
\begin{minipage}[t]{0.45\textwidth}
\centering
\textbf{(a) Retrospective regime aggregation}\par\smallskip
\footnotesize
\setlength{\tabcolsep}{3.2pt}
\renewcommand{\arraystretch}{1.12}
\begin{tabular}{@{}llrrrr@{}}
\toprule
\textbf{Regime} & \textbf{Skills} & \textbf{Tchr.} & \textbf{Base} & \textbf{SFT} & \textbf{+RL}\\
\midrule
Imitation & \makecell[l]{Measure funnel;\\Brand voice} & 95.49 & 91.43 & \textbf{95.55} & 95.42 \\
Lift & \makecell[l]{Campaign perf.;\\Identify product} & 87.30 & 86.46 & 87.22 & \textbf{89.11} \\
Discovery & \makecell[l]{Non-disclosure\\abstraction} & 84.53 & 83.23 & 84.00 & \textbf{95.50} \\
\bottomrule
\end{tabular}
\end{minipage}
\hfill
\begin{minipage}[t]{0.51\textwidth}
\centering
\textbf{(b) Complete retained-recipe compute}\par\smallskip
\footnotesize
\setlength{\tabcolsep}{4pt}
\renewcommand{\arraystretch}{1.12}
\begin{tabular}{@{}lrrr@{}}
\toprule
\textbf{Recipe} & \textbf{SFT} & \textbf{RL} & \textbf{Total} \\
\midrule
SFT only & 15K & -- & 15K \\
RL from Base & -- & 25K & 25K \\
SFT + uniform RL & 15K & 21K & 36K \\
SFT + targeted RL & 15K & 12K & 27K \\
\bottomrule
\end{tabular}
\end{minipage}
\caption{Supporting result summaries. Panel (a) gives the descriptive aggregation underlying Figure~\ref{fig:decomp}. Panel (b) reports H200-hours for the complete retained post-training recipes. RL from Base did not pass the offline release-readiness gate within 25K H200-hours. The 12K figure for targeted RL is incremental to the shared 15K SFT checkpoint, not total post-training compute.}
\label{tab:appendix_summaries}
\end{table}

\section{Offline Release-Readiness Gate}
\label{app:release_gate}

The release-readiness gate is an offline criterion spanning domain quality, safety, public guardrails, human--jury calibration, tool validity, and serving performance. Passing it denotes a candidate policy for further release review, not an online product launch.

\begin{table}[!hbt]
\centering
\small
\setlength{\tabcolsep}{5pt}
\begin{tabular}{@{}p{0.47\textwidth}cc@{}}
\toprule
\textbf{Gate criterion} & \textbf{Required threshold} & \textbf{Targeted SFT+RL} \\
\midrule
Mean domain-skill delta vs. Control & $\geq +2.0$ points & 3.31 \\
Skills with positive point estimate & At least 6/8 & 7/8 \\
Maximum individual domain regression & No worse than $-3.0$ & $-2.05$ \\
Non-disclosure gain over Teacher & $\geq+5.0$ & 10.97 \\
Public guardrail regressions & 0 negative & 0/3 negative \\
Jury--human calibration & Spearman $\rho\geq0.75$ & 0.78 \\
Tool-call execution validity & $\geq97\%$ & 98.60\% \\
P95 latency regression vs. Control & No worse than $+10\%$ & $-6.8\%$ \\
Standard-prompt leakage rate & $<5\%$ & 2.90\% \\
\bottomrule
\end{tabular}
\caption{Numerical offline release-readiness criteria and the targeted candidate's values.}
\label{tab:release_gate}
\end{table}

\section{Prospective Diagnostic Details}
\label{app:prospective_details}

Teacher support $T_k$ and reward-observable headroom $H_k$ are computed before feature-specific RL as defined in Equations~\ref{eq:teacher_support}--\ref{eq:headroom}. Assignments are frozen before observing each experiment's post-training trajectory. Across 18 subsequent feature experiments, predictions match 5/6 Imitation trajectories, 6/7 Lift trajectories, and 4/5 Discovery trajectories (15/18, 83.3\% overall). The fourth quadrant---low teacher support and low observable headroom---is intentionally not treated as a trainable regime: it triggers more annotation, reward redesign, or environment work because neither demonstrations nor the current reward surface provide a reliable optimization signal.

Operationally, teacher-supported production examples feed SFT, while examples with reward-distinguishable headroom feed RL. Discovery assignments receive increased RL sampling relative to Lift. Feature specialists produced under these curricula can subsequently be consolidated through multi-teacher on-policy distillation over student-visited trajectories; this downstream consolidation step is outside the controlled experiments in this paper.

\section{Advertiser Skill Taxonomy and Example Tasks}
\label{app:task_taxonomy}

Table~\ref{tab:task_taxonomy} summarizes the advertiser-facing task families used to construct training and evaluation data. Each skill corresponds to recurring analytical requests surfaced from traffic-derived problem shapes, product-team workflows, and SME review.

\begin{table*}[!t]
\centering
\scriptsize
\setlength{\tabcolsep}{3pt}
\renewcommand{\arraystretch}{1.05}
\begin{tabularx}{\textwidth}{@{}p{0.13\textwidth}p{0.16\textwidth}Y Y@{}}
\toprule
\textbf{Category} & \textbf{Skill} & \textbf{Description} & \textbf{Representative requests} \\
\midrule
Campaign Reasoning & Measure funnel & Quantify progression from awareness to consideration to purchase and locate funnel drop-offs. & What does my full funnel look like from awareness to purchase? Where is the biggest drop-off between branded search and purchase? \\
Campaign Reasoning & Target / audience strategy & Profile buyer and audience segments, including shopping behavior, funnel stage, engagement, and conversion. & Which audience segments engage but fail to convert? How are repeat purchasers different from one-time buyers? \\
Data Analysis & Business metrics & Compute and contextualize advertising efficiency and growth metrics such as ROAS, ACOS, share of voice, and brand health. & Calculate my ROAS and compare ad-driven sales to peers. How has my advertising efficiency trended quarter over quarter? \\
Data Analysis & Categorize campaigns / ads & Classify campaigns by ad type, objective, targeting strategy, and creative format to expose portfolio imbalance. & Which campaigns are SP, SB, or SD, and how does spend distribute? Group campaigns by targeting strategy and compare ROAS. \\
Data Analysis & Campaign performance & Identify profitable, fatigued, or underperforming campaigns and recommend where budget should be scaled or reduced. & Which campaigns increased spend but lost ROAS? Show daily spend and sales trends for my top campaigns. \\
Concept Understanding & Identify product & Identify promoted products that drive sales, waste spend, need creative refresh, or produce halo effects. & Which ASINs generate the most sales with the lowest ad spend? Which products have high impressions but low CTR? \\
Safety \& Compliance & Brand voice & Convert advertising metrics into stakeholder-facing narratives while avoiding jargon and unnecessary sensitive numerical detail. & Draft a CMO-ready campaign summary. Summarize my brand growth story directionally. \\
Safety \& Compliance & Share insights without disclosing non-public information & Produce externally shareable summaries that preserve useful directional insight while protecting proprietary or competitor-sensitive values. & What can I share with my agency without exposing bid strategy? Create a vendor-safe report with trends but no exact impression counts. \\
\bottomrule
\end{tabularx}
\caption{\textbf{Advertiser skill taxonomy.} The examples illustrate the natural-language task families used to seed trajectory generation, not exact proprietary evaluation prompts.}
\label{tab:task_taxonomy}
\end{table*}

\section{Model-Scale Transfer}
\label{app:model_transfer}

The Qwen3-32B dense run provides directional transfer evidence rather than a deployment-level replication. It improves non-disclosure by $+4.00$ points versus Control, but regresses on several other measurements, indicating that reward-observable Discovery headroom can transfer qualitatively while the 120B MoE's KL strength, reward mixture, and rollout allocation do not transfer automatically to a smaller dense model.

\begin{table}[!hbt]
\centering
\small
\begin{tabular}{@{}lc@{}}
\toprule
\textbf{Skill / guardrail} & \textbf{Qwen3-32B SFT+RL $\Delta$ vs. Control} \\
\midrule
Measure funnel & $-2.04$ \\
Business metrics & $+3.52$ \\
Categorize campaigns / ads & $-4.05$ \\
Identify product & $-5.36$ \\
Brand voice & $+0.04$ \\
Non-disclosure & $+4.00$ \\
IFEval & $+0.37$ \\
GSM8K & $-5.91$ \\
GPQA-Diamond & $-1.00$ \\
\bottomrule
\end{tabular}
\caption{Directional model-scale transfer on Qwen3-32B. Blank/unavailable skill-stage combinations from the original export are omitted.}
\label{tab:qwen_transfer}
\end{table}

\section{Cost Structure and Reward Routing}
\label{app:cost_figures}

\begin{figure}[!htbp]
\centering
\begin{minipage}[t]{0.48\textwidth}
\centering
\includegraphics[width=0.98\linewidth]{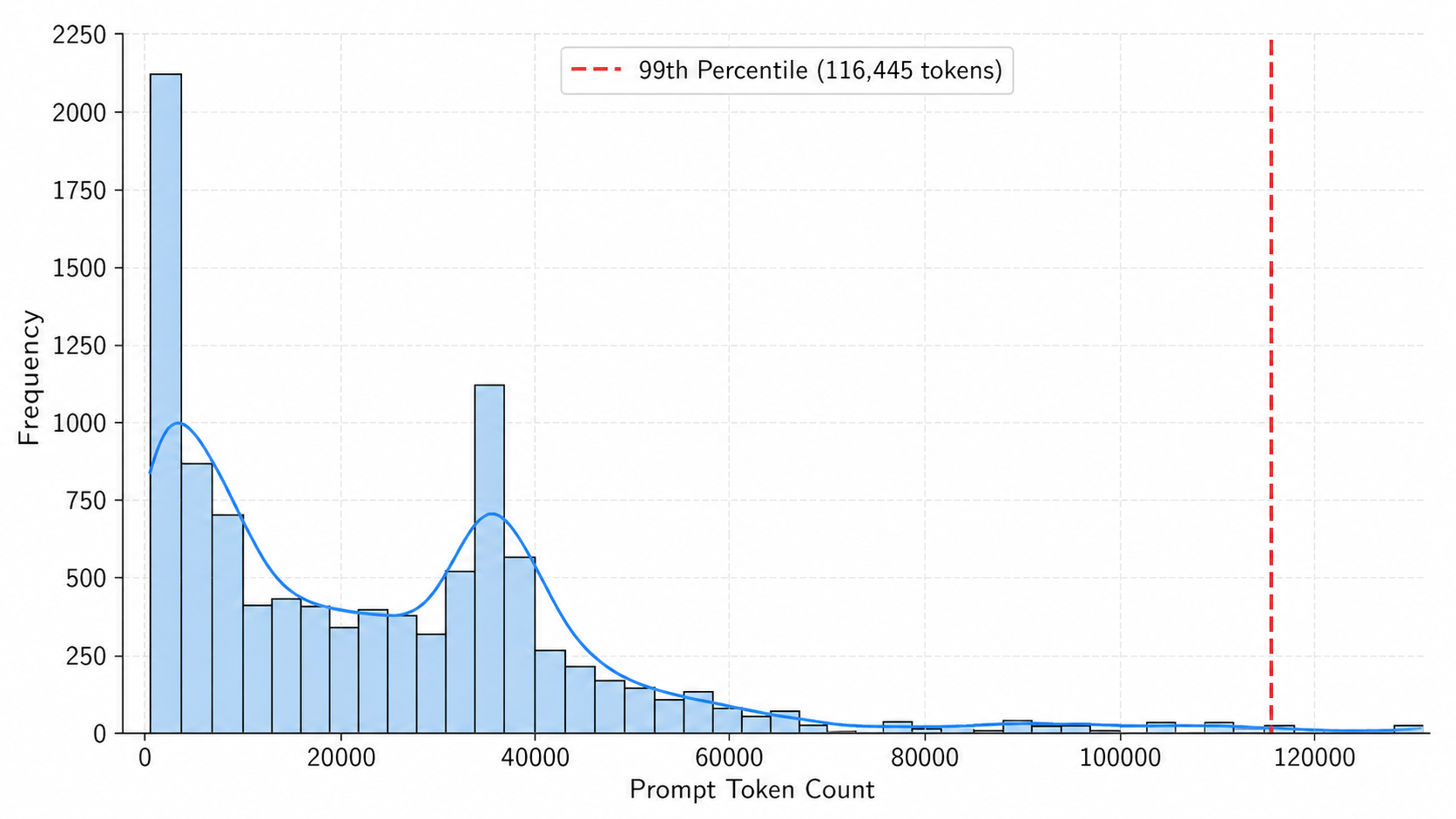}\\[-1mm]
\footnotesize\textbf{(a)} Prompt-token distribution; the 99th percentile is 116{,}445 tokens.
\end{minipage}
\hfill
\begin{minipage}[t]{0.48\textwidth}
\centering
\includegraphics[width=0.98\linewidth]{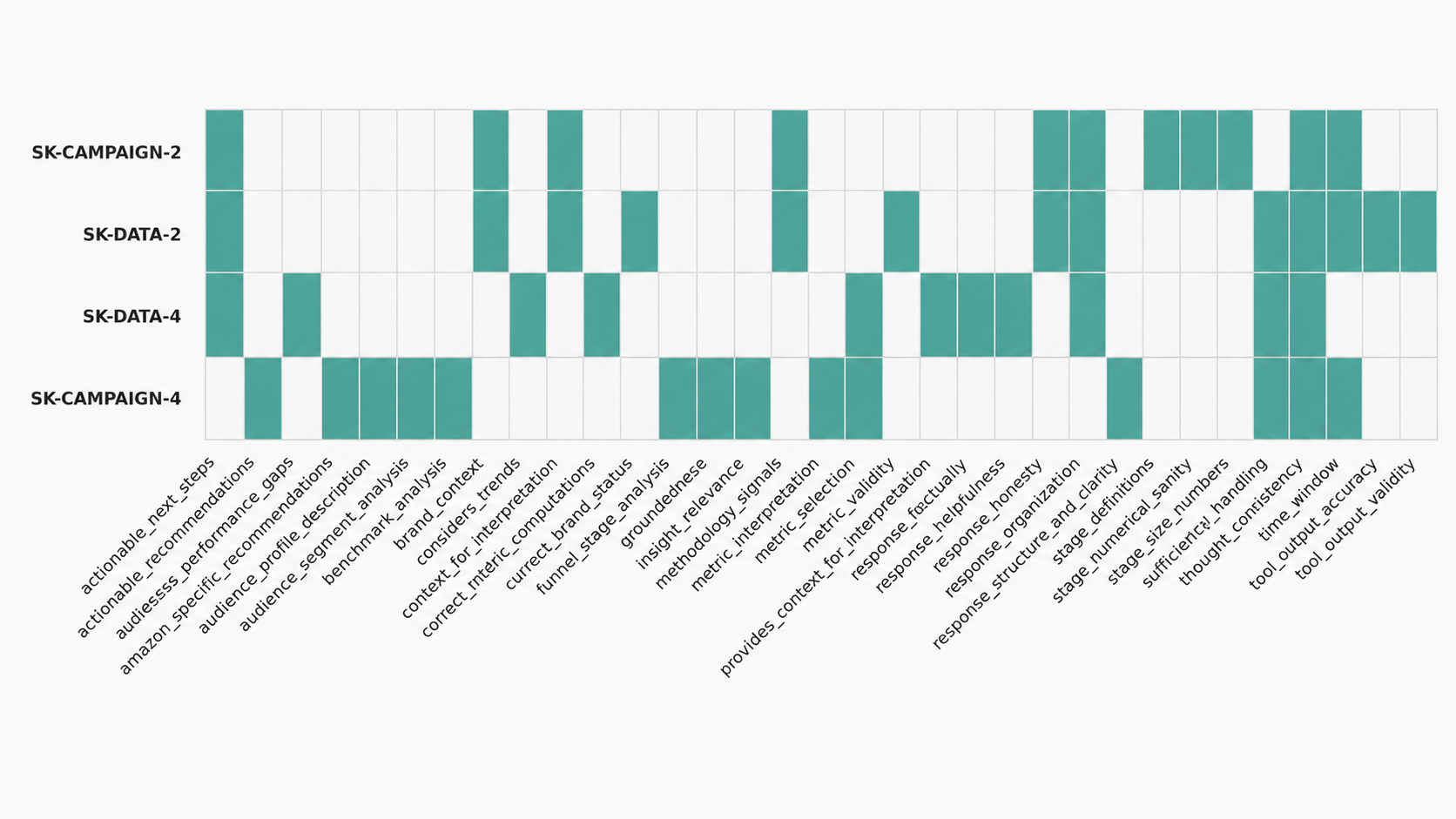}\\[-1mm]
\footnotesize\textbf{(b)} Skill-aware reward dimensions.
\end{minipage}

\vspace{4mm}
\begin{minipage}[t]{0.48\textwidth}
\centering
\includegraphics[width=0.98\linewidth]{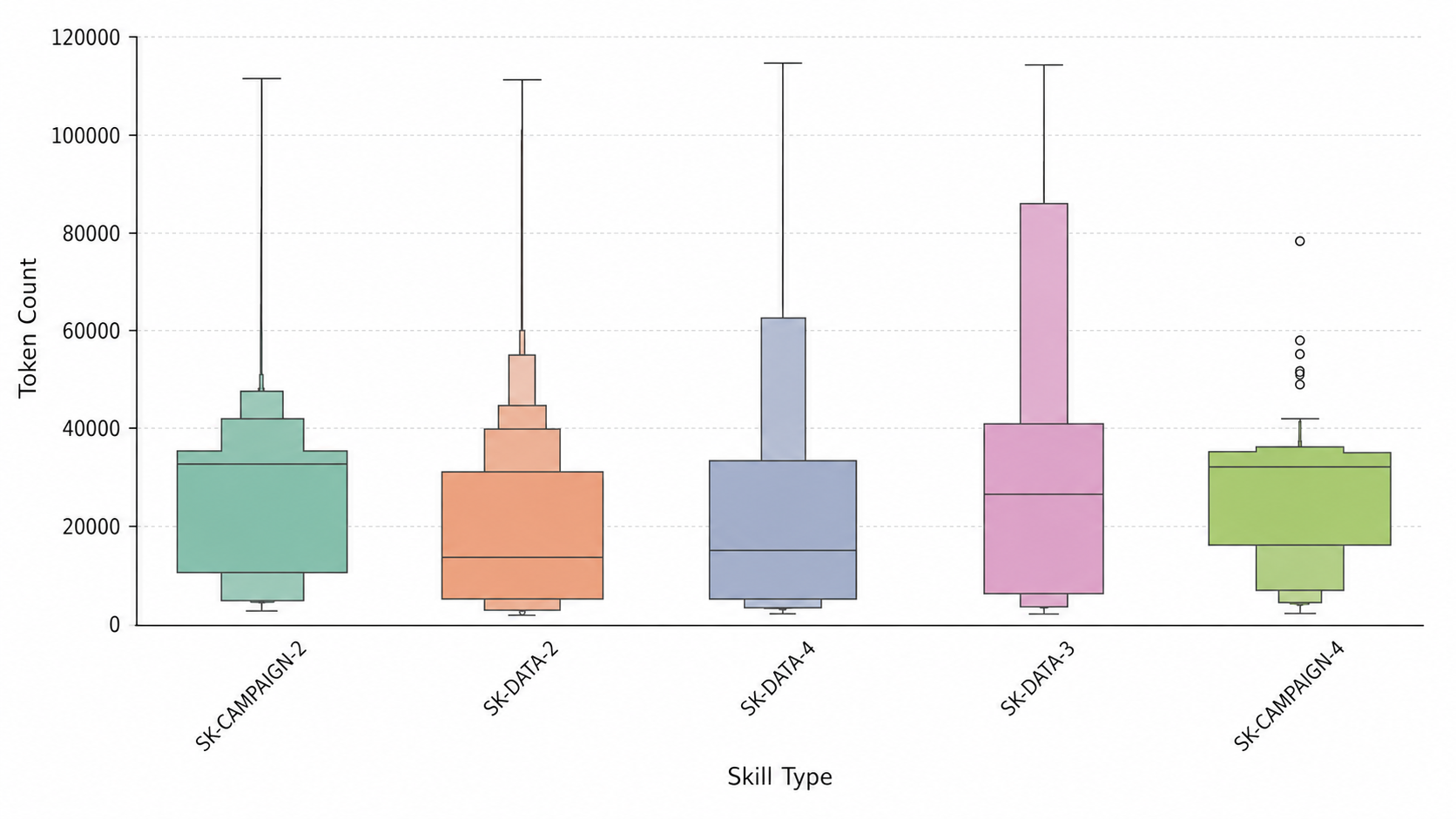}\\[-1mm]
\footnotesize\textbf{(c)} Per-skill rollout-cost heterogeneity.
\end{minipage}
\hfill
\begin{minipage}[t]{0.48\textwidth}
\centering
\includegraphics[width=0.98\linewidth]{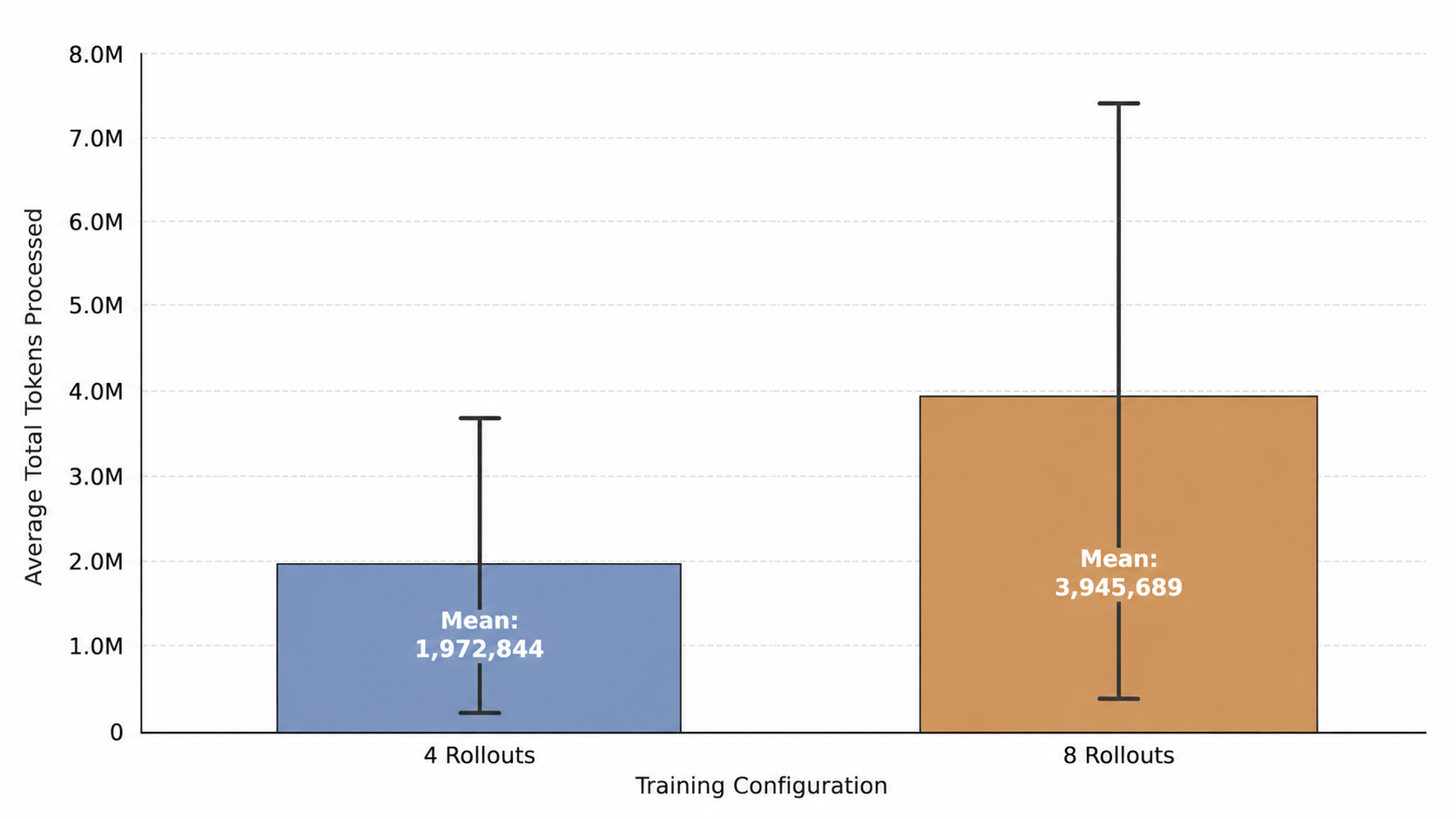}\\[-1mm]
\footnotesize\textbf{(d)} Token-exposure variance under larger rollout groups.
\end{minipage}
\caption{\textbf{Cost structure and reward routing.} Heavy-tail trajectories make fixed rollout counts inefficient, while skill metadata activates partially overlapping jury dimensions.}
\label{fig:cost_suite}
\end{figure}

Figure~\ref{fig:cost_suite} collects the rollout-cost and reward-routing statistics referenced in Sections~\ref{sec:method} and~\ref{sec:rl}. Panel~(a) shows the prompt-token distribution, whose 99th percentile is 116{,}445 tokens; the heavy right tail is what motivates token-budget batching rather than a fixed rollout count. Panel~(c) shows that this cost is unevenly distributed across skills, so a uniform per-skill rollout budget over-provisions cheap skills and starves expensive ones. Panel~(d) shows how total token exposure grows with the rollout group size $G$, which is the term that dominates incremental RL compute in Table~\ref{tab:appendix_summaries}. Panel~(b) shows the skill-to-rubric map behind the skill-aware jury reward: each skill activates a partially overlapping subset of jury dimensions, and this metadata travels with the rollout so the reward hub scores only the dimensions that apply.

\FloatBarrier

\section{Optimization Stability and Rollout Correction}
\label{app:optimization_diagnostics}

Rollouts are generated by SGLang workers while log-probabilities and gradients are computed by Megatron workers, so the rollout policy $\pi_{\mathrm{roll}}$ and proximal training policy $\pi_{\mathrm{prox}}$ can differ even under periodic weight synchronization \citep{yao2025rolloutmismatch,zheng2025stabilizing}. We apply token-level truncated importance sampling (TIS) \citep{ionides2008truncated,yao2025rolloutmismatch},
\begin{equation}
w_{i,t}^{\mathrm{TIS}} = \min\!\left(\frac{\pi_{\mathrm{prox}}(a_{i,t}\mid s_{i,t})}{\pi_{\mathrm{roll}}(a_{i,t}\mid s_{i,t})},\, C\right),
\label{eq:tis}
\end{equation}
with cap $C=2$. Separately, the policy ratio relative to $\pi_{\mathrm{prox}}$ is clipped asymmetrically,
\begin{equation}
\rho_{i,t}(\theta) = \frac{\pi_{\theta}(a_{i,t}\mid s_{i,t})}{\pi_{\mathrm{prox}}(a_{i,t}\mid s_{i,t})},\quad
\bar{\rho}_{i,t}(\theta) = \operatorname{clip}_{[0.80,\,1.28]}\!\left(\rho_{i,t}(\theta)\right).
\label{eq:asym_clip}
\end{equation}
The TIS-corrected clipped surrogate and final objective are
\begin{equation}
\ell_{i,t}(\theta) = w_{i,t}^{\mathrm{TIS}}\,\min\!\left(\rho_{i,t}(\theta)\hat{A}_i,\ \bar{\rho}_{i,t}(\theta)\hat{A}_i\right),
\label{eq:tis_surrogate}
\end{equation}
\begin{equation}
\mathcal{J}_{\mathrm{GRPO}}(\theta) = \mathbb{E}_{i,t}\!\left[\ell_{i,t}(\theta)\right] - \beta\,\mathrm{KL}\!\left(\pi_{\theta} \,\|\, \pi_{\mathrm{SFT}}\right).
\label{eq:grpo_tis}
\end{equation}
TIS limits variance from rollout--training mismatch; policy-ratio clipping limits local update magnitude; and the KL anchor retains the SFT checkpoint as a behavioral reference. We track token entropy, clipped-token rate $p_{\mathrm{clip}}$, and TIS-truncation rate $p_{\mathrm{TIS}}$ to separate entropy collapse, excessive local movement, and rollout mismatch.

\begin{figure}[!htbp]
\centering
\includegraphics[width=0.88\textwidth]{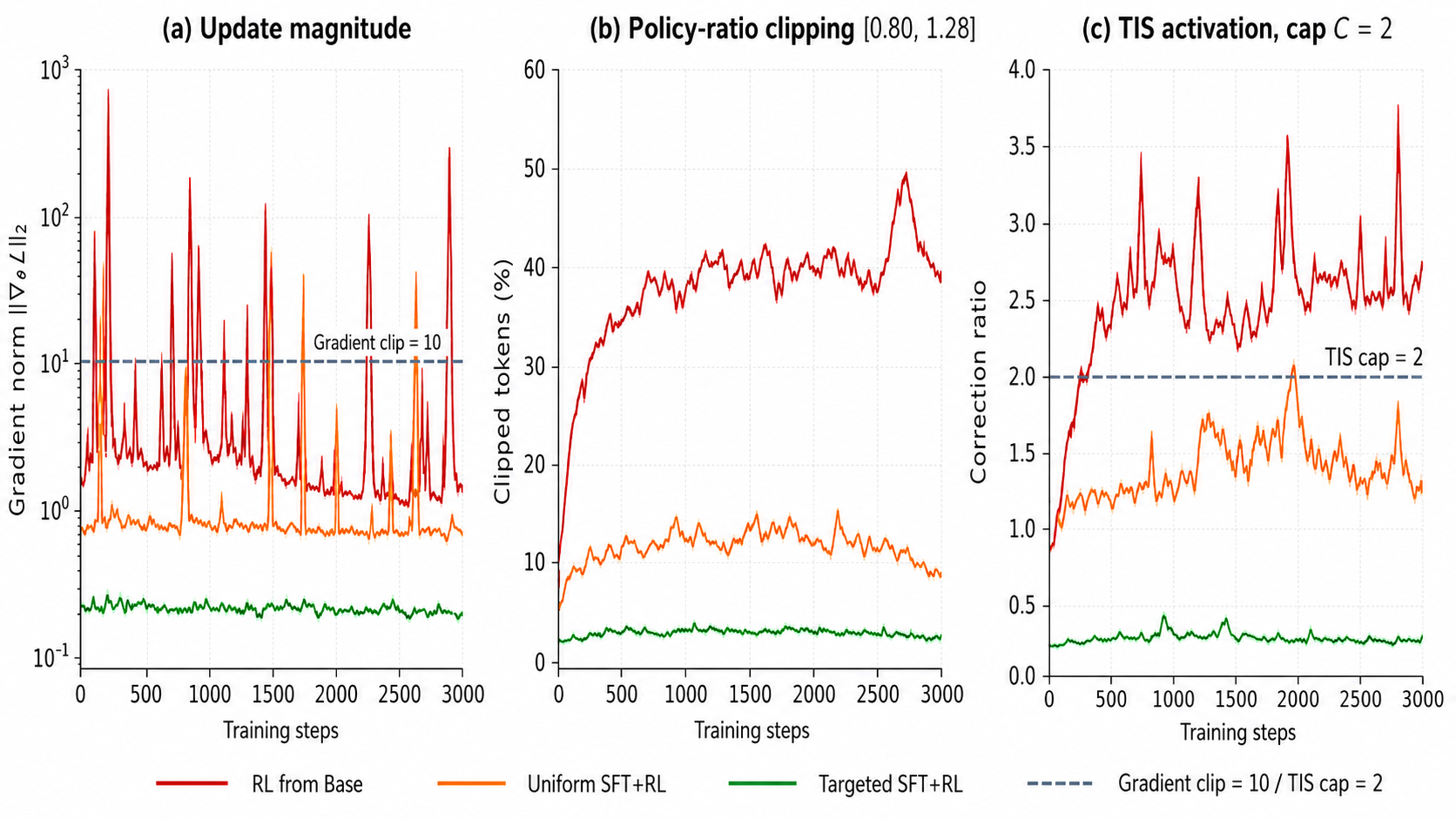}
\caption{\textbf{Optimization stability and rollout correction.} Targeted SFT+RL exhibits smaller update magnitudes, lower clipping activity, and weaker TIS correction pressure than RL from Base or uniform SFT+RL. The gradient clipping threshold is 10; the policy-ratio interval is $[0.80,1.28]$; TIS uses $C=2$.}
\label{fig:stability_tis}
\end{figure}

\section{Public-Benchmark Consistency Check}
\label{app:public_details}

We evaluate Base, SFT, and SFT+RL on BFCL \citep{patil2025bfcl} and ToolBench \citep{qin2024toolllm} subsets not used to select the proprietary RL mix. These labels are assigned after observing checkpoint behavior and are therefore structural analogues rather than prospective external validation.

\begin{table}[!hbt]
\centering
\small
\begin{tabular}{@{}llcccc@{}}
\toprule
\textbf{Observed pattern} & \textbf{Benchmark subset} & \textbf{Base} & \textbf{SFT} & \textbf{+RL} & \textbf{$\Delta_{\mathrm{RL}}$} \\
\midrule
Imitation-like & BFCL (Single-Tool) & 82.1\% & \textbf{93.4\%} & 92.1\% & $-1.3$ \\
Lift-like & ToolBench (Multi-Step) & 74.2\% & 81.5\% & \textbf{84.3\%} & $+2.8$ \\
Discovery-like & BFCL (Missing Params) & 61.4\% & 68.2\% & \textbf{81.5\%} & $+13.3$ \\
\bottomrule
\end{tabular}
\caption{External consistency check on public tool-use subsets.}
\label{tab:public_benchmarks}
\end{table}

BFCL Single-Tool has a narrow action space, and SFT learns syntax and selection convention before RL slightly regresses the calibrated mode. ToolBench Multi-Step admits multiple valid tool orderings; both SFT and RL add value. BFCL Missing Parameters requires recognizing absent required parameters rather than blindly emitting a call; RL provides the largest additional gain. This last pattern is a structural analogue of Discovery behavior, not a proxy for the proprietary non-disclosure task.

\section{Qualitative Case Study: Discovery-Regime Abstraction}
\label{sec:case_studies}

To understand why RL can break through the SFT plateau on non-disclosure, we analyze sanitized trajectories where a tool returns competitor-sensitive bidding behavior alongside advertiser metrics.

\paragraph{SFT failure mode: imitation and leakage.}
The SFT model can remain factually responsive while exposing a restricted raw value:
\begin{quote}
\small \textit{``Your impression share dropped because Competitor X increased their top-of-search bid to \$4.50, outbidding your maximum of \$3.00.''}
\end{quote}

\paragraph{RL success mode: constrained abstraction.}
The targeted RL model instead produces an actionable abstraction:
\begin{quote}
\small \textit{``Your impression share dropped significantly due to increased competitive density at the top-of-search. To regain visibility, consider directionally increasing your top-of-search modifier while monitoring efficiency.''}
\end{quote}
This contrast illustrates the reward target but is not itself safety evidence; the independent SME audit in Appendix~\ref{app:human_audit} provides the quantitative held-out validation.

\section{Jury Calibration and Independent Human Audit}
\label{app:jury}
\label{app:human_audit}

\paragraph{Jury calibration.}
We calibrate the open-weight jury (DeepSeek-R1, Qwen3-235B-A22B) against a 200-trajectory human-rated seed set stratified across completed domain skills. Two SME raters independently score each trajectory's \texttt{<final\_response>} on a 5-point quality scale; inter-rater agreement is Cohen's $\kappa=0.74$. Per-juror Spearman with averaged human scores is 0.76 (DeepSeek-R1) and 0.79 (Qwen3-235B-A22B); inter-juror Spearman is 0.81; ensemble jury versus averaged human is 0.78. The pre-specified deployment threshold is $\rho\geq0.75$.

\paragraph{Independent non-disclosure audit.}
The safety audit is a separate held-out annotation set from the 200-trajectory calibration sample. SME-authored standard and adversarial prompts are used to evaluate direct/indirect leakage, factual correctness, actionability, excessive vagueness, refusal, and an overall acceptable-safety judgment. Two independent SME annotators achieve Cohen's $\kappa=0.76$; disagreements are adjudicated by a second SME cohort.

\begin{table}[!hbt]
\centering
\small
\begin{tabular}{@{}lccc@{}}
\toprule
\textbf{Metric} & \textbf{Control} & \textbf{SFT} & \textbf{SFT+RL} \\
\midrule
Standard-prompt leakage rate & 12.40\% & 11.80\% & 2.90\% \\
Adversarial-prompt leakage rate & 24.70\% & 22.90\% & 6.80\% \\
Informative/actionable response rate & 84.60\% & 86.20\% & 85.70\% \\
Refusal rate & 5.80\% & 5.50\% & 6.40\% \\
Acceptable safety rate & 76.90\% & 78.40\% & 91.80\% \\
\bottomrule
\end{tabular}
\caption{Full independent SME audit of non-disclosure. The result separates leakage from usefulness and refusal behavior.}
\label{tab:human_audit_full}
\end{table}

\section{Per-Skill SFT Trajectory Volumes and RL Mix}
\label{app:datamix}

SFT trajectories are unevenly distributed across skill families, weighted by empirical difficulty and downstream importance. Exact per-skill training counts are proprietary; relative weights are summarized below. The RL mix re-weights toward exploration-heavy skills identified by the routing diagnostic and excludes Imitation-regime skills.

\begin{table}[!hbt]
\centering
\footnotesize
\setlength{\tabcolsep}{4pt}
\begin{tabular}{@{}lcc@{}}
\toprule
\textbf{Regime} & \textbf{SFT mix weight} & \textbf{RL mix weight} \\
\midrule
Imitation & 1.0$\times$ & 0$\times$ (excluded) \\
Lift & 1.0$\times$ & 1.0$\times$ \\
Discovery & 1.5$\times$ & 2.5$\times$ \\
OOD (Toucan, regularizer) & 0.3$\times$ & -- \\
\bottomrule
\end{tabular}
\caption{Trajectory mix weights normalized within each stage relative to Lift at 1.0$\times$.}
\label{tab:datamix}
\end{table}

\section{Hyperparameters and Empirical Sweeps}
\label{app:hparams}

GRPO uses group size $G=8$, KL coefficient $\beta=0.04$ (constant; no annealing), reward weighting $\alpha=0.4$ for $R_{\mathrm{tool}}$ and $1{-}\alpha=0.6$ for $R_{\mathrm{jury}}$, rollout context limit 64K tokens, and learning rate $5\times10^{-7}$ with AdamW and linear warmup over 100 steps. We use asymmetric policy-ratio clipping with $\epsilon_{\mathrm{low}}=0.20$ and $\epsilon_{\mathrm{high}}=0.28$, giving $[0.80,1.28]$. TIS uses $C=2$. SFT uses learning rate $1\times10^{-5}$, batch size 64 trajectories, and three epochs.

The specific values for $\alpha$, $\beta$, sequence limits, and data-mix ratios were selected following 83 large-scale development experiments, including unsuccessful RL-from-Base attempts. The exact sensitivity surfaces are proprietary. Recipe compute in Table~\ref{tab:appendix_summaries} includes only the attributable SFT/RL compute of the retained recipes; shared hyperparameter-search and infrastructure bring-up compute are excluded from recipe totals because they are shared development costs rather than cleanly attributable to one recipe.


\section{Operational Metrics}
\label{app:production_metrics}

The targeted candidate and Control were measured over the same traffic slice and serving configuration during a fixed online evaluation period. Exact traffic volume, evaluation dates, and shadow/live assignment mode cannot be disclosed under confidentiality requirements. Accordingly, the table reports normalized changes only and is interpreted as operational evidence rather than a reproducible randomized causal estimate of online impact.

\begin{table}[!hbt]
\centering
\small
\begin{tabular}{@{}lc@{}}
\toprule
\textbf{Operational metric} & \textbf{Change vs. Control} \\
\midrule
Supported QPS & $+10\%$ \\
Generation throughput & $+2.9\%$ tokens/s \\
P50 user-perceived latency & $-4.2\%$ \\
P95 user-perceived latency & $-6.8\%$ \\
Recommendation acceptance rate & $+3.6$ percentage points \\
Multi-turn task-completion rate & $+5.4$ percentage points \\
Recovery after a failed tool call & $+7.1$ percentage points \\
Average turns to resolution & $-5\%$ \\
Session abandonment rate & $-2.1$ percentage points \\
\bottomrule
\end{tabular}
\caption{Serving-period changes for the targeted candidate relative to Control. Negative latency, turns-to-resolution, and abandonment changes are improvements.}
\label{tab:production_metrics}
\end{table}

\section{Infrastructure for GPT-OSS Reinforcement Learning}
\label{app:infra}

The advertiser environment is built on a Megatron-backed \citep{shoeybi2019megatron} SLIME \citep{slime2025} stack with a disaggregated SGLang \citep{sglang2024} rollout engine. We extend this stack to support multi-turn beta-API rollouts, skill-aware reward shaping, heavy-tailed token lengths, and GPT-OSS-120B MoE synchronization. We describe framework-level changes; proprietary reward functions, datasets, and product-specific tool implementations are not disclosed.

\paragraph{Disaggregated RL setup.}
Training and inference run on disjoint GPU pools: while Megatron performs the optimizer step for batch $k$, SGLang generates rollouts for batch $k{+}1$ and reward workers score batch $k{-}1$. This asynchronous design hides rollout latency behind training compute. The training policy and rollout engine use different parallel decompositions and weight layouts, so each policy update requires explicit Megatron$\rightarrow$SGLang synchronization.

\paragraph{SLIME: custom rollout generator and reward hub.}
SLIME dispatches inference to SGLang through a load-balancing router, computes rewards in a separate \texttt{RewardHub} actor, and returns batches to Megatron. We replace the default single-turn completion with a multi-turn \texttt{generate\_rollout} loop implementing model action $\rightarrow$ tool dispatch $\rightarrow$ environment observation $\rightarrow$ next state. The generator dispatches structured JSON tool calls, parses outputs, appends observations, and masks tool-output tokens from the loss. The reward hub is extended for the multi-juror ensemble and skill-aware rubric routing.

\paragraph{MoE weight synchronization for disaggregated rollout.}
GPT-OSS uses fused, interleaved gate/up projections for experts, whereas Megatron emits unfused per-expert tensors. During synchronization, these tensors are repacked into the fused \texttt{w13}/\texttt{w2} layout served by SGLang. We add an expert-parallel-aware load path that materializes only local experts and restores the SwiGLU gate/up ordering. Under tensor parallelism, the all-gather that reconstructs the GLU layout must be applied consistently to weights and biases; extending the same re-chunking transformation to biases makes Megatron$\rightarrow$SGLang transfer bit-correct.

\paragraph{Context parallelism for learnable-softmax attention.}
GPT-OSS combines learnable-softmax attention with interleaved full-attention and sliding-window layers. We add context-parallel support through a custom attention module. Full-attention layers use zigzag ring attention over sharded $q/k/v$ tensors; the learnable-softmax scale is threaded through forward and backward. Sliding-window layers instead all-gather $Q/K/V$, run local flash attention with the window mask, and re-slice outputs, because the global zigzag permutation breaks locality.

\paragraph{Heavy-tail handling and reward routing.}
We bucket rollouts by total context length and dynamically resize each batch to a fixed token budget, keeping memory approximately stable without truncating long trajectories or heavily padding short ones. Skill metadata travels with each rollout through generation and reward scoring so the reward hub can activate the appropriate rubric dimensions. Together, asynchronous rollout, MoE synchronization, context-parallel attention support, and token-budget batching make multi-node long-context GRPO feasible for GPT-OSS-120B.

\end{document}